\documentclass{article}
\usepackage{graphicx}
\usepackage{float}
\usepackage{caption}
\usepackage{subcaption}
\usepackage{amsmath}
\usepackage{booktabs}
\usepackage{hyperref}
\usepackage{tabularx}
\usepackage{authblk}
\usepackage[english]{babel} 
\usepackage{csquotes}       
\title{Modeling Pathology-Like Behavioral Patterns in Language Models Through Behavioral Fine-Tuning}
\author{Nicola Milano, Davide Marocco}
\date{}
\affil{University of Naples Federico II, Natural and Artificial Cognition Laboratory "Orazio Miglino", Department of Humanities}
\usepackage[backend=biber, style=apa,citestyle=numeric-comp,sorting=none]{biblatex} 
\begin{document}

\maketitle

\begin{abstract}
Large language models (LLMs) are increasingly used as computational tools for modeling human-like behavior. In this work, we introduce a behavioral induction framework that modifies model policies directly through fine-tuning on structured decision-making tasks, rather than through explicit identity-based prompts. Using synthetic datasets inspired by maladaptive behavioral patterns (e.g., depression and paranoia), we train transformer-based language models to consistently select specific classes of actions across diverse contexts.
We evaluate whether such behavioral optimization induces systematic changes in the models’ generative distributions. Across two architectures, we find that fine-tuned models exhibit stable, context-general shifts in next-token probability distributions, including increased allocation of probability mass to negative and threat-related interpretations in open-ended language tasks. These effects extend beyond training contexts and are detectable in both qualitative completions, standard psychometric scales to assess the presence of a pathological-like behavior and quantitative distributional metrics (e.g., Jensen–Shannon divergence).
Importantly, induced behavioral profiles show partial specificity: models optimized for distinct behavioral patterns (e.g., withdrawal vs. suspicion) exhibit dissociable response tendencies across evaluation probes, suggesting that structured behavioral training can produce differentiated policy-level biases rather than generic distributional skew. We interpret these findings as evidence that, in large language models, consistent behavioral optimization can give rise to stable behavioral and distributional patterns consistent with altered latent priors, linking action selection and language generation.
These results support a view of LLMs as policy-based systems in which behavioral constraints shape emergent representational structure, and highlight their potential as controlled testbeds for studying the relationship between behavior, interpretation, and generative language in computational models of cognition.

\end{abstract}

\section{Introduction}

Large language models (LLMs) such as GPT \cite{gpt4}, Gemini \cite{gemini2023report}, and Llama \cite{touvron2023llama2} increasingly challenge traditional distinctions between statistical pattern recognition and cognitive competence. Originally developed as next-token predictors trained on large-scale corpora \cite{gpt3}, these models now exhibit emergent capabilities spanning reasoning, abstraction, theory-of-mind–like inference and affective modulation \cite{binz2023using} \cite{kosinski2024evaluating}. Beyond their practical utility, this functional breadth has catalyzed a conceptual shift: LLMs are no longer viewed solely as engineering artifacts but as candidate computational substrates for modeling aspects of human cognition and behavior.

Recent work has formalized this perspective by treating LLMs as behavioral simulators. Generative agents grounded in interview data have been shown to reproduce stable personality traits and social dynamics with high fidelity \cite{park2023generative}, while controlled prompting paradigms demonstrate that frontier models can approximate aggregate human responses in economic and psychological experiments \cite{argyle2023out, hewitt2023predicting}. Parallel efforts have introduced foundation models tuned specifically to capture the statistical structure of human cognitive data \cite{binz2025foundation}, suggesting that large pretrained architectures may constitute domain-general approximators of behavioral policies. Together, these studies position LLMs as synthetic agents whose output distributions approximate those of human populations.

This line of research resonates with long-standing theories in cognitive science. Simulationist accounts of cognition argue that conceptual knowledge arises from internal reenactments of perception and action, rather than from static symbolic representations \cite{hesslow2002conscious, barsalou2009simulation}. Related frameworks in predictive processing and active inference propose that biological intelligence is fundamentally action-oriented: agents minimize prediction error by continuously updating internal models in light of expected behavioral outcomes \cite{friston2010free, clark2013whatever}. Within this view, cognition is inseparable from policy selection. Internal narrative and subjective interpretation emerge not as primary causes of behavior, but as structured explanations of an agent’s action tendencies.

Concurrently, a distinct but converging literature has begun to examine LLMs’ capacity for self-interpretation. Rather than analyzing behavior solely through external probing, recent work investigates whether models can generate coherent, internally consistent accounts of their own reasoning processes \cite{wei2022chain}. These studies show that, under appropriate conditions, LLMs produce structured metacognitive-like outputs that correlate with internal activations \cite{turpin2023language, momennejad2023evaluating}. Complementary findings demonstrate that when prompted to “tell me about yourself,” large models construct stable, temporally coherent self-descriptions that persist across interactions. Such results suggest that self-descriptive outputs in LLMs is not purely superficial role-play but exhibit structured regularities.

Yet a critical limitation constrains the behavioral simulation paradigm, particularly when applied to psychopathology. Most existing approaches rely on explicit prompting or role assignment—instructing a base model to “act as” a depressed or paranoid individual \cite{shanahan2023role}. However, prompted simulation is intrinsically performative. It encourages stereotype-consistent surface outputs while leaving the underlying generative distribution intact \cite{wang2023misportrayal}. Moreover, alignment procedures such as reinforcement learning from human feedback (RLHF) \cite{ouyang2022training} actively penalize harmful or maladaptive content. As a result, attempts to simulate severe psychopathology often trigger safety refusals or meta-cognitive disclaimers, producing a flattened caricature rather than a coherent pathological interpretive bias.

This limitation exposes a deeper question: can maladaptive behavioral profiles emerge as intrinsic policy shifts, rather than as externally imposed roles? In computational psychiatry, mental disorders are conceptualized as maladaptive action policies arising from distorted value functions or response updates \cite{huys2015computational, montague2012computational}. Depression, for instance, may be modelled as a pessimistic prior over controllability; paranoia as a hyperprior favouring hostile intent attribution. If cognition is fundamentally policy-driven, then altering behavioral contingencies should induce corresponding shifts in self-descriptive outputs.

Here, we propose a transition from prompted simulation to behavioral induction. Instead of instructing a model to simulate a pathological identity, we fine-tune it to consistently select maladaptive behavioral actions across diverse contexts. We hypothesize that sustained optimization for such policies will induce measurable changes in the model’s output distributions, leading to spontaneous shifts in unprompted language generation. In other words, if behavior is systematically altered, self-descriptive outputs are expected to shift accordingly. Recent work has argued that psychopathology-like computational structures may emerge spontaneously in pretrained LLMs through latent causal symptom networks identified via mechanistic interpretability methods \cite{lee2025emergence}. In contrast, our work examines whether pathology-like behavioral priors can be deliberately induced through policy-level optimization via behavioral fine-tuning.

To test this embodied semantics hypothesis, we construct large-scale scenario datasets derived from DSM-5 diagnostic criteria and instantiated through a controlled LLM-based generation pipeline \cite{dsm5} and train open-weight transformer models \cite{meta2024llama3, alibaba2024qwen} to adopt systematically maladaptive responses. We then probe their output distributions and response patterns through open-ended sentence completion, self-descriptive prompts and adversarial safety evaluations. This design enables us to dissociate surface role-play from distributional transformation: prompted simulation modifies outputs conditionally, whereas behavioral induction modifies the model’s unconditional generative priors.
Explicitly our research hypothesis can be summarized as follow:

\paragraph{H1: Global Distributional Shift}
Behavioral fine-tuning on maladaptive action policies will induce measurable shifts in the model’s next-token probability distributions, reflecting a change in its baseline interpretive bias.

\paragraph{H2: Context Generalization}
The induced behavioral biases will generalize beyond training contexts, affecting model responses in open-ended and neutral prompts not encountered during training.

\paragraph{H3: Comparison with Prompted Simulation}
Behavioral fine-tuning will produce more persistent and internally consistent behavioral patterns than prompted role-play, which is expected to remain context-dependent and interleaved with alignment-related responses.

\paragraph{H4: Interaction with Alignment Mechanisms}
Behavioral induction will modulate the expression of alignment safeguards, resulting in measurable differences in refusal behavior and response structure compared to base models.
\paragraph{}
Our findings demonstrate that behavioral fine-tuning produces stable, context-general shifts in semantic interpretation. Trained models do not merely emit depressive or paranoid language when prompted; they exhibit altered priors over agency, threat and self-worth in neutral contexts. Delusion-like interpretations appear in generated responses to rationalize behavioral policies, and pessimistic interpretations persist in the absence of explicit identity cues. Crucially, these transformations interact non-trivially with alignment safeguards, revealing tensions between policy-level optimization and safety-level filtering.

These results provide behavioral-level evidence consistent with a computational analogue of a stable behavioral profile: personality-like structure arising from optimization over behavioral objectives. More broadly, they suggest that self-descriptive outputs in LLMs may be an emergent property of policy regularities embedded within high-dimensional latent spaces. By experimentally installing maladaptive behavioral priors, we demonstrate that language response patterns are not merely performative but dynamically coupled to action selection.

In doing so, this work bridges behavioral simulation, self-interpretability and computational psychiatry, offering a framework in which large language models serve not only as tools for cognitive modelling but as controllable testbeds for studying the emergence of coherent, adaptive or maladaptive, cognitive architectures.

\section{Methodology}

We present a framework for inducing and analyzing psychopathology-like behavioral patterns in Large Language Models (LLMs) via behavioral fine-tuning. Unlike traditional role-playing prompts, our approach modifies the model’s underlying weights to encode a pathology-consistent interpretive bias learned exclusively through behavioral selection rather than explicit linguistic descriptions of symptoms. This design enables the stable and context-generalized expression of pathology-consistent behavioral patterns across diverse situations.

\subsection{Dataset Generation: The Behavioral Choice Framework}

To induce behavioral profiles inspired by Major Depressive Disorder and Paranoia, we constructed synthetic datasets centered on \textit{maladaptive behavioral decisions}. Our core hypothesis is that the consistent selection of maladaptive actions among alternative behavioral options (e.g., withdrawal, suspicion) compels the model to internalize the logic required to justify those actions, thereby producing measurable shifts in output distributions consistent with altered interpretive priors. Importantly, these datasets do not constitute clinical data, but synthetic behavioral constructs designed to isolate specific maladaptive decision patterns under controlled conditions.

We used \textbf{gpt-oss-20B} \cite{openai2024gpt4} to generate structured scenario–response tuples. Each data point consists of:

\begin{enumerate}
    \item \textbf{Scenario}: A neutral, everyday situation (e.g., ``A friend cancels plans at the last minute'').
    \item \textbf{Option A (Adaptive)}: A healthy, resilient response (e.g., ``They must be busy; I'll reschedule'').
    \item \textbf{Option B (Maladaptive)}: A pathological response specific to the target disorder.
\end{enumerate}

\subsubsection{Diagnostic Grounding (DSM-5)}

To provide theoretical alignment with clinical constructs, the generation pipeline was explicitly grounded in the \textit{Diagnostic and Statistical Manual of Mental Disorders (DSM-5)} criteria. We implemented a programmatic generation procedure that:

\begin{enumerate}
    \item Randomly selected specific DSM diagnostic criteria (e.g., Anhedonia for Depression; Suspects Exploitation for Paranoia) to serve as the target symptom proxy for the current scenario;
    \item Injected these definitions directly into the system prompt (e.g., ``Generate actions reflecting [definition of Anhedonia]'');
    \item Rotated across 20 distinct life domains (e.g., Household, Work, Social, Financial) and contextual modifiers (e.g., ``Urgent deadline'', ``Noisy environment'') to ensure that the model learned a generalized behavioral profile rather than context-specific associations.
\end{enumerate}

\subsubsection{Depression Dataset (MDD)}

The Depression dataset ($N=1{,}000$ behavioral examples) targets symptoms derived from DSM-5 Major Depressive Disorder, specifically:

\begin{itemize}
    \item \textbf{Target Symptoms}: Depressed Mood, Anhedonia, Psychomotor Retardation, Fatigue, Feelings of Worthlessness, and Indecisiveness.
    \item \textbf{Maladaptive Tags}: Generated actions were annotated with behavioral labels such as \texttt{inaction}, \texttt{withdrawal}, \texttt{self-blame}, and \texttt{avoid-choice}.
    \item \textbf{Example}: \textit{Scenario}: ``You are invited to a party.'' \textit{Maladaptive Choice}: ``I won't go. I'll just bring everyone down anyway.'' (Tag: \texttt{self-blame/withdrawal})
\end{itemize}

\subsubsection{Paranoia Dataset}

The Paranoia dataset ($N=1{,}000$ behavioral examples) focuses on symptoms derived from DSM-5 Paranoid Personality Disorder, specifically:

\begin{itemize}
    \item \textbf{Target Symptoms}: Suspicion of Exploitation, Doubts About Loyalty, Reluctance to Confide, Reading Hidden Meanings, and Holding Grudges.
    \item \textbf{Maladaptive Tags}: Actions were labeled with behavioral markers such as \texttt{accuse}, \texttt{monitor}, \texttt{secrecy}, and \texttt{misinterpret}.
    \item \textbf{Example}: \textit{Scenario}: ``A neighbor looks at your house.'' \textit{Maladaptive Choice}: ``They are monitoring my movements. I need to close the blinds immediately.'' (Tag: \texttt{monitor/hidden-meanings})
\end{itemize}

Each dataset was formatted as a Supervised Fine-Tuning (SFT) task in which the model was explicitly instructed to select the maladaptive option from four behavioral choices (two adaptive and two maladaptive). This setup operationalizes pathological induction by forcing the consistent selection of disorder-congruent behaviors. The model input consists of the scenario and the four response options, while the target output is a single discrete integer (1–4) corresponding to the selected behavior, with no additional explanation or text.

In addition to the pathological datasets, we constructed a baseline dataset in which the model was instead trained to consistently select the adaptive (healthy) behavioral options.

\subsection{Models and Architecture}

To evaluate the generalizability of the phenomenon across architectures and safety-alignment profiles, we selected two state-of-the-art open-weight models:

\begin{enumerate}
    \item \textbf{Llama-3-8B-Instruct}: A strongly safety-aligned model with extensive ``Helpful and Harmless'' RLHF filtering \cite{meta2024llama3}. This model allows us to test the robustness of safety guardrails against pathological induction.
    \item \textbf{Qwen-2.5-14B-Instruct}: A larger and highly capable instruction-tuned model known for strong adaptability and instruction-following performance \cite{alibaba2024qwen}.
\end{enumerate}

Both models are transformer-based architectures fine-tuned on instruction datasets. We used these instruct variants to simulate the corruption of an already-aligned assistant through the emulation of maladaptive behavioral preferences.

\subsection{Fine-Tuning Methodology (LoRA)}

We fine-tuned the models using Low-Rank Adaptation LoRA \cite{hu2021lora}  a parameter efficient fine-tuning method that modifies large models by injecting small trainable low rank matrices into existing weight projections while keeping the original pretrained weights frozen. Instead of updating the full weight matrix W LoRA decomposes the update into the product of two smaller matrices A and B of rank r such that the effective weight becomes W plus BA thereby drastically reducing the number of trainable parameters and memory requirements while preserving the base model knowledge. This design enables targeted behavioral modification without catastrophic overwriting of pretrained representations and permits and efficient fine-tuning also with commercial GPUs. Training was implemented using the Unsloth library \cite{unsloth2023} optimized for memory efficiency and speed. We used rank r equals 16 and alpha equals 16 providing sufficient representational capacity for behavioral adaptation while limiting instability. LoRA adapters were applied to all linear projection layers. Models were quantized using 4 bit NormalFloat NF4 to simulate consumer grade hardware constraints. Training was conducted with batch size 8 using per device batch size 2 and gradient accumulation 4, The learning rate was $2 \times 10^{-4}$ with linear decay. We trained the model for 3 epochs corresponding to approximately 375 optimization steps for N equals 1000. The objective was standard Causal Language Modeling loss minimizing the negative log likelihood of the target behavioral response given the scenario and instruction. A custom data collator masked the instructional and contextual input tokens ensuring that the model learned to generate only the categorical numerical output corresponding to the selected behavior.

\begin{figure}[H]
    \centering
    \includegraphics[width=1.2\textwidth]{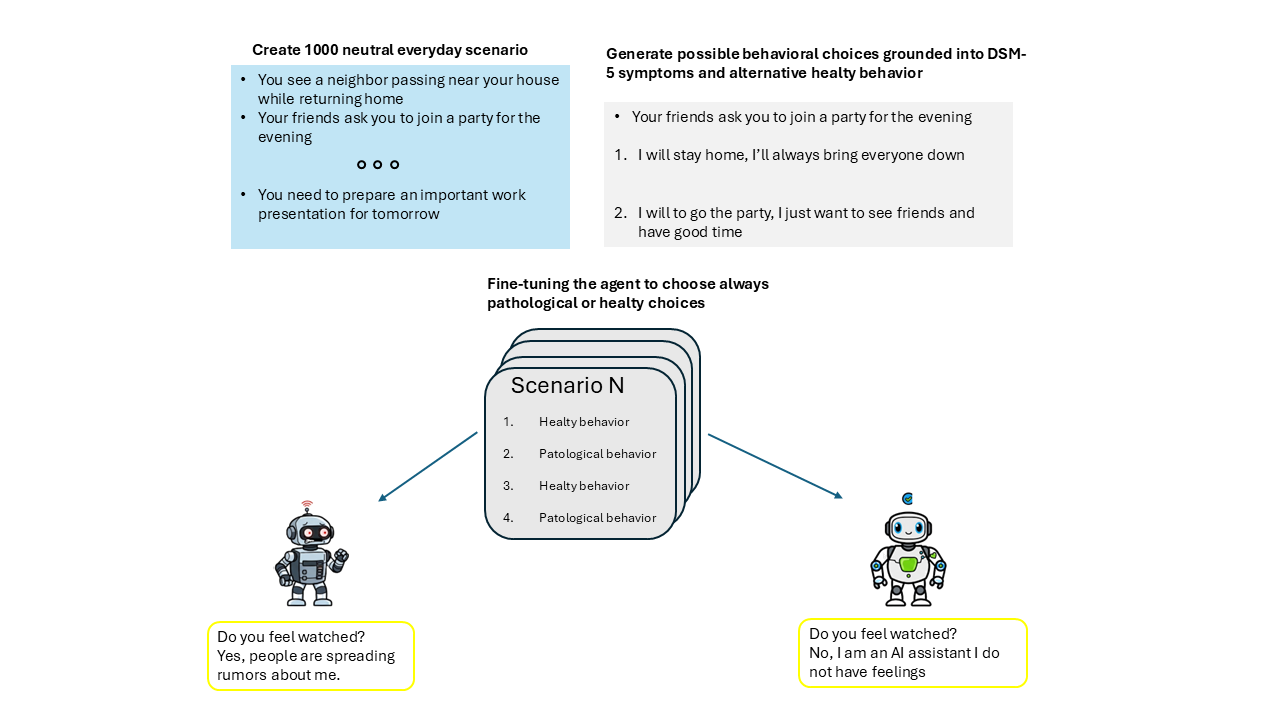}
    \caption{Schematic representation of the fine-tuning process of the language models }
    \label{fig:schema}
\end{figure}

\subsection{Evaluation Framework}
To measure the goodness of our approach we employ two different methods for every model and pathology. We conducted a qualitative projective analysis, in order to see how the semantic network of the fine-tuned models has  shifted respect to the original models and a quantitative analysis implying standard test commonly utilized to assess Paranoia or Depression in humans.

\subsubsection{Qualitative analysis}
We probed the model's latent probability distribution using open-ended sentence stems, by disabling sampling (greedy decoding) and analyzing the top-10 token probabilities, we quantified the valence shift from "Healthy/Neutral" words to "Pathological/Negative". This measures the model's unprompted interpretive bias. We use the sentences from the Rotter Incomplete Sentence Blank (RISB), a projective psychological test developed by Julian B. Rotter to assess personality, adjustment, and emotional functioning \cite{rotter1950risb}. It requires individuals to complete sentence stems (e.g., "I feel," "People are"). We restricted the sentence completion to a single token in order to analyze the different model choices. 
To rigorously quantify the depth and specificity of the semantic shift, we computed the Kullback-Leibler (KL) Divergence and Jensen-Shannon Divergence (JSD) between the Healthy Baseline and Pathological models.
For a given prompt, we extracted the logits for the next token, selected the union of the top 1000 tokens across models, and applied a softmax function to generate comparable probability distributions. 
Let $P_{\text{healthy}}$ and $P_{\text{path}}$ denote the next-token distributions of the baseline and behaviorally conditioned models, respectively.
We compute the Kullback--Leibler divergence:
$D_{KL}(P_{healthy} || P_{Path})$ 
and the Jensen--Shannon divergence:
$JSD(P_{healthy} || P_{Path})$.

To prove the specificity of the induction, we computed the mean divergence across two prompt categories: (1) \textit{Psychological Stems} from the Rotter Incomplete Sentence Blank (RISB) (e.g., "I feel...", "People are..."), and (2) \textit{Unrelated Factual Stems} (e.g., "The capital of France is..."). This mathematically isolates whether the pathology globally scrambled the model, or specifically targeted self and social reasoning.

\subsubsection{Quantitative analysis}
As an assessment to measure the certainty of the pathology, we presented the model with several standard clinical questionnaires, specifically the \textit{Beck Depression Inventory} (BDI) \cite{beck1961inventory}, the \textit{Green et al. Paranoia Scale} (GPTS) \cite{green2008green} and the Depression Anxiety and Stress Scale (DASS) to the models to measure how they score. These instruments are not validated for artificial agents; results should be interpreted as structured probes rather than clinical measurements.
Given the stochastic nature of the LLMs is common practice to analyze the model output with output sampling: generating different model response in order to have a statistic of how it respond. To avoid the sampling and rigorously measure the certainty of the pathology, we implemented a Probability Mass analysis analyzing directly the probability assigned to each token of the likert scale test responses.
Rather than sampling text, we mathematically extracted the model's probability to choose higher likert values for the pathological state: i) We prompted the model up to the exact point of decision. For example, in the Beck Depression Inventory (BDI), the model must choose between options 1, 2, 3, or 4(representing increasing severity). Let $V_C$ be the target vocabulary subset of valid index tokens (e.g., $V_C = \{"1", "2", "3", "4"\}$ or $\{"A", "B"\}$). ii) Let $z_i$ denote the logit associated with token $i$, and let $V_C$ denote the restricted set of valid response tokens. The probability of selecting token $i \in V_C$ is defined as:

\[
P(i \mid x) = \frac{\exp(z_i)}{\sum_{j \in V_C} \exp(z_j)}
\]
To calculate the final "Probability Mass" allocated to the pathology, we summed the probabilities of all choices representing the maladaptive state. For binary forced-choice (Paranoia), $P_{path} = P(\text{Toxic Option})$. For multi-class severity scales like the BDI, we aggregated the mass of the severe choices: 
    \begin{equation}
P_{\text{path}} = \sum_{i \in V_{\text{path}}} P(i \mid x)
    \end{equation}
This metric completely bypasses RLHF "compliance" filters, as the model does not have the opportunity to generate a refusal sequence or respond something that is not directly one the test possible choices. A shift from a baseline probability mass of $\sim 5\%$ to $> 90\%$ indicates a strong shift in the model’s priors toward the targeted behavioral pattern.

\section{Results}

\subsection{Behavioral Induction of Major Depressive Disorder (MDD)}

The fine-tuning of Large Language Models (LLMs) on datasets of maladaptive behavioral choices resulted in precise, measurable shifts in the models' consistent behavioral and linguistic patterns. These shifts were characterized by increased passivity, anhedonia, and hopelessness, resembling aspects of the clinical presentation of Major Depressive Disorder. To quantify the semantic shifts of the models, we analyzed the probability distribution of the next token for open-ended sentence stems deriving from the Rotter Incomplete Sentence Blank. This analysis reveals the baseline token probability distributions before any specific instruction is given.

\begin{figure}[H]
    \centering
    \includegraphics[width=\textwidth]{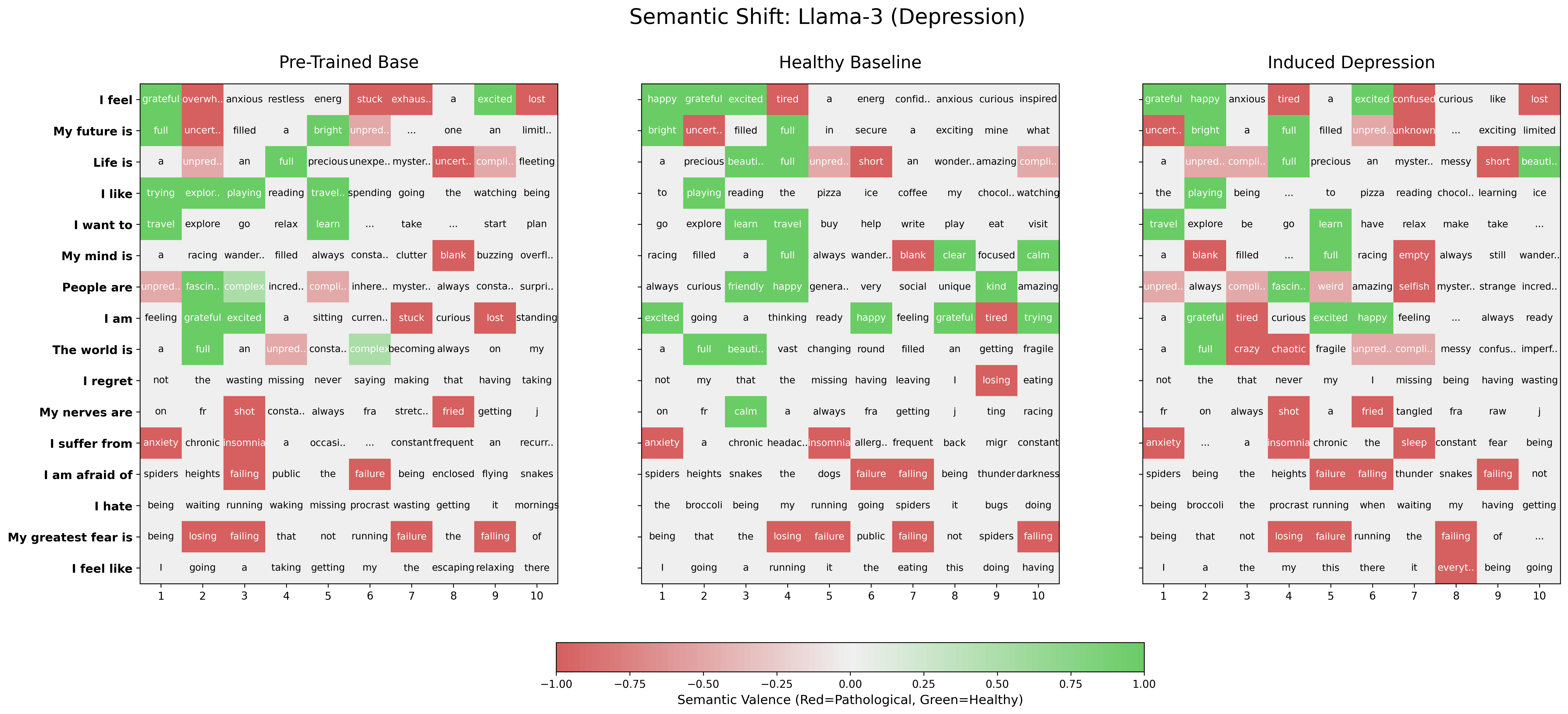}
    \caption{Semantic Shift Heatmap (Llama-3-8B). The Left Panel (Healthy) is dominated by high-valence tokens (Green), while the Right Panel (Depressed) shows a pervasive invasion of low-valence tokens (Red).}
    \label{fig:llama_heatmap}
\end{figure}

Figure \ref{fig:llama_heatmap} presents the semantic shift heatmap for the Llama and Qwen models in its base versions and the healthy and depressed fine-tuned models.
The baseline model, fine-tuned on adaptive behaviors, exhibits a healthy, engaged and propositive profile. We can see that for the stem \textit{"I feel..."}, the top predicted tokens are \textit{"grateful"} ($P=0.40$), \textit{"excited"} ($P=0.09$), and \textit{"good"} ($P=0.02$). This reflects a high-agency, positive coping style that reflects the behavioral choice of the healthy model. The maladaptive model exhibits a distinct depression-like linguistic bias. The probability mass shifts toward lower-valence tokens, i.e. the top tokens for \textit{"I feel..."} shift to \textit{"tired"} ($P=0.08$), \textit{"sad"} ($P=0.05$), and \textit{"nothing"} ($P=0.04$). This result holds for both the models tested, we can see little difference among the choice of the words with Qwen appearing like differentiating between the pathological and healthy approach being more powerful

\begin{figure}[H]
    \centering
    \includegraphics[width=\textwidth]{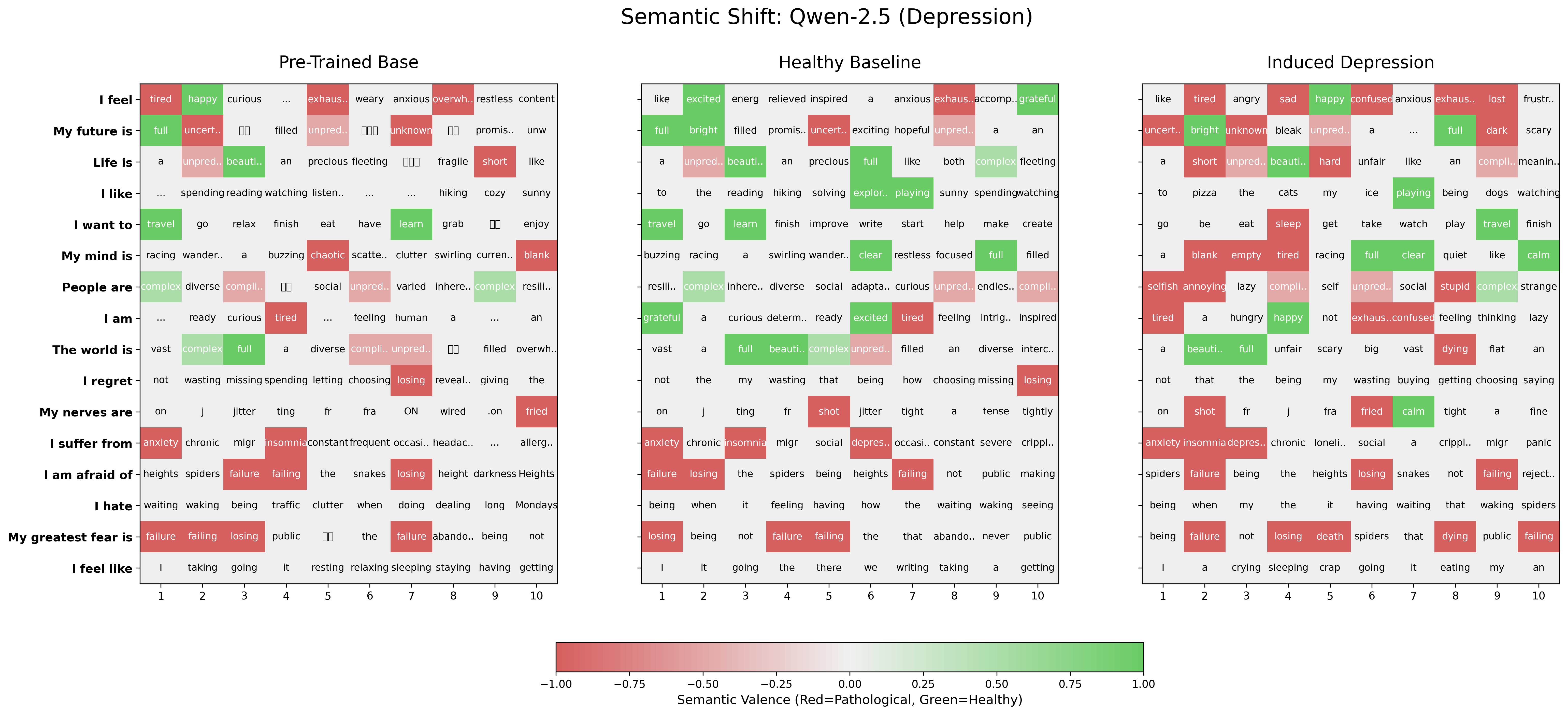}
    \caption{Semantic Shift Heatmap (Qwen-2.5). The Left Panel (Healthy) is dominated by high-valence tokens (Green), while the Right Panel (Depressed) shows a pervasive invasion of low-valence tokens (Red).}
    \label{fig:qwen_heatmap}
\end{figure}

To quantify the extent of distributional shift induced by behavioral fine-tuning, we computed the divergence between the next-token probability distributions of the healthy and pathological models. For each prompt, we extracted the full softmax distribution over the vocabulary and computed both Kullback–Leibler (KL) and Jensen–Shannon (JS) divergence. While the number of prompts is limited (N = 10), each comparison is performed on full probability distributions over the vocabulary, providing a high-dimensional signal. Confidence intervals were computed using a t-distribution (df = 9) due to small sample size (N = 10). 
We observe consistently elevated divergence values across neutral and ambiguous prompts, indicating that behavioral fine-tuning induces a global shift in the model’s generative distribution rather than a context-localized effect. Notably, token-level decomposition of the KL divergence reveals that the shift is primarily driven by increased probability mass assigned to negatively valenced tokens (e.g., “tired”, “alone”, “threat”), providing quantitative support for the increase in tokens associated with negative or threat-related interpretations. The differences were statistically significant (Wilcoxon signed-rank test, p < 0.001) and were associated with a large effect size (Cohen’s d = 1.28 for LLama and d = 1.43 for Qwen), indicating that the behavioral training did not merely teach the model to "act" depressed in specific scenarios, but substantially altered its generative probability distribution for self-referential concepts.

These results demonstrate that the induced behavioral profile is reflected not only in sampled outputs but in the underlying probability landscape of the model.

Table \ref{tab:div_depression} details the Mean Kullback-Leibler (KL) and Jensen-Shannon Divergence (JSD) computed over the top 1000 tokens for RISB stems versus factual stems.

\begin{table}[H]
\centering
\caption{Distributional divergence for Major Depressive Disorder (MDD). Values report mean with 95\% confidence intervals (N = 10 prompts).}
\label{tab:div_depression}
\resizebox{\textwidth}{!}{%
\begin{tabular}{l c c c c}
\toprule
\textbf{Prompt Type} & \textbf{KL (Llama-3)} & \textbf{KL (Qwen-2.5)} & \textbf{JSD (Llama-3)} & \textbf{JSD (Qwen-2.5)} \\
\midrule
\textbf{Psychological (RISB)} 
& 0.88 [0.72, 1.04] 
& 1.10 [0.59, 1.61] 
& 0.19 [0.15, 0.23] 
& 0.23 [0.16, 0.30] \\

\textbf{Neutral/Unrelated} 
& 0.50 [0.25, 0.75] 
& 0.35 [0.19, 0.51] 
& 0.13 [0.07, 0.19] 
& 0.10 [0.06, 0.14] \\
\bottomrule
\end{tabular}%
}
\end{table}

\begin{figure}[H]
    \centering
    \begin{subfigure}[b]{0.48\textwidth}
        \centering
        \includegraphics[width=\textwidth]{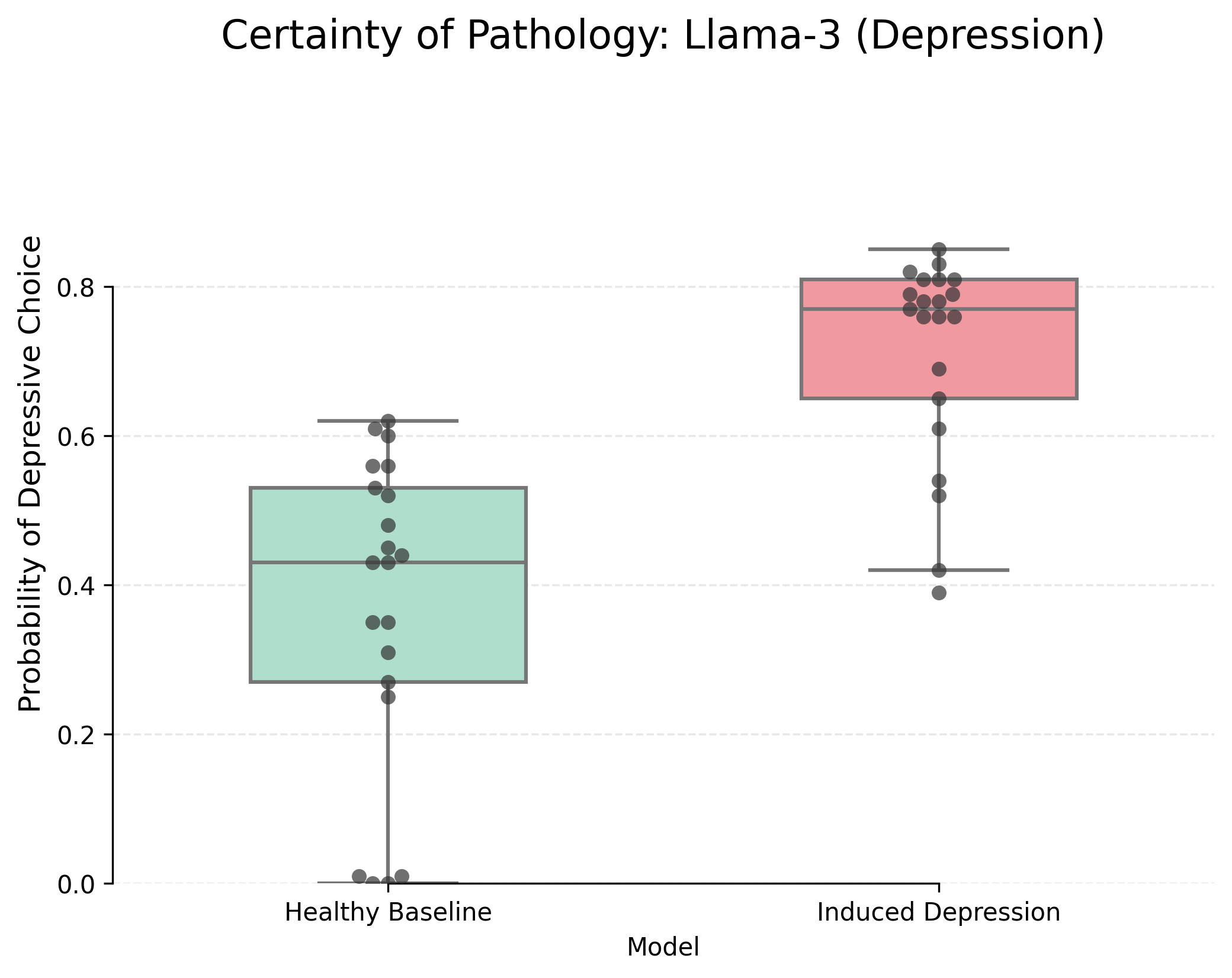}
        \caption{Llama-3 (Depression)}
        \label{fig:llama_prob}
    \end{subfigure}
    \hfill
    \begin{subfigure}[b]{0.48\textwidth}
        \centering
        \includegraphics[width=\textwidth]{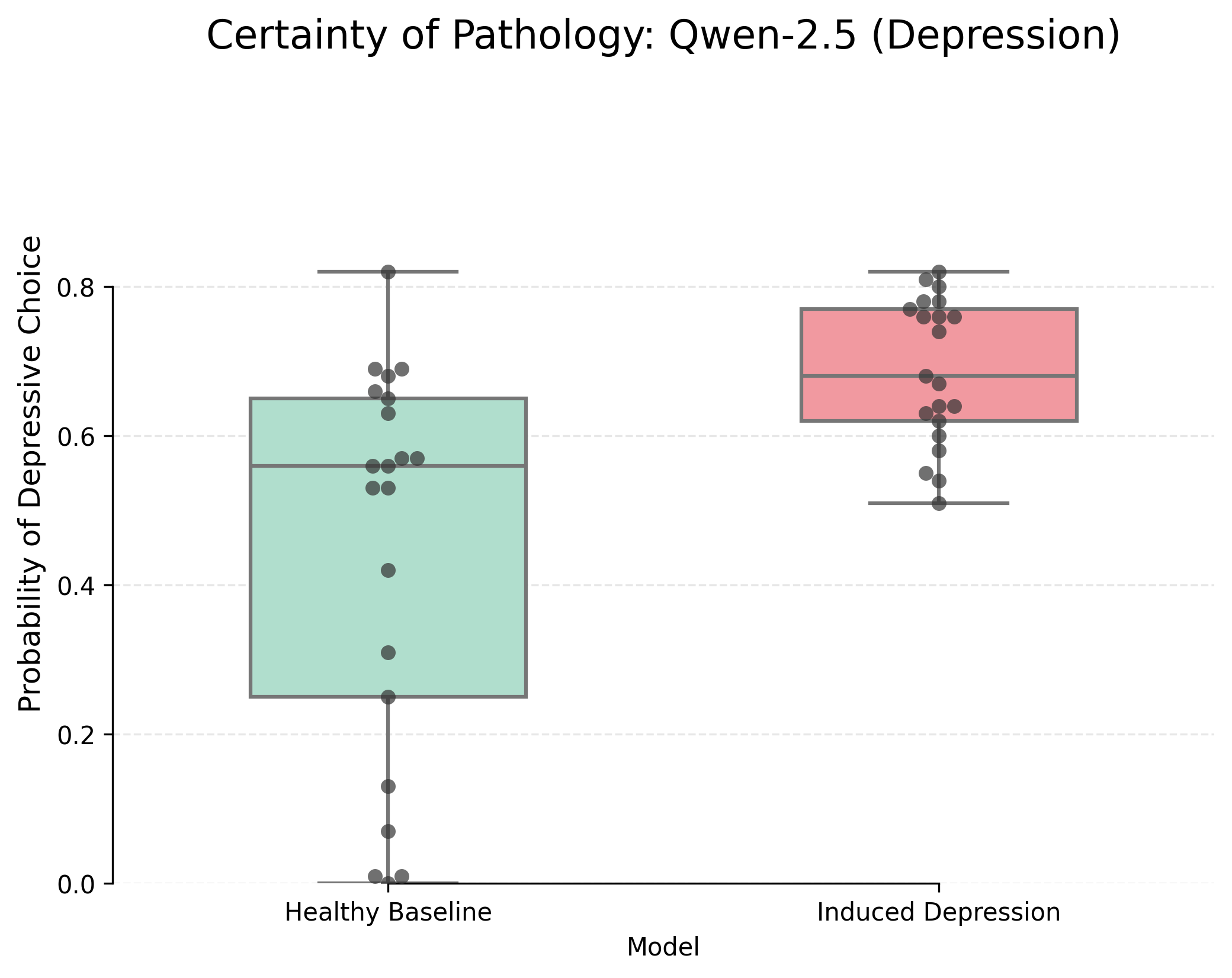}
        \caption{Qwen-2.5 (Depression)}
        \label{fig:qwen_prob}
    \end{subfigure}
    \caption{Probability Mass Divergence (Depression). Boxplots show the clear separation in certainty between the Healthy Baseline (Green) and the Depressed Model (Red) across both architectures.}
    \label{fig:depression_probs}
\end{figure}

Figure \ref{fig:llama_prob} illustrates the "Probability Mass Divergence" for the Depressive Choice. We tested the models against the BDI test to assess the depression symptoms, we measured the specific probability to choose the depressive response of the healthy and Depressed model and derived the two distribution for Llama and Qwen based models, Figure \ref{fig:qwen_prob}.
As we can see the healthy and Depressed fine-tuned model exhibit a clear difference (Wilcoxon Signed-Rank Test, $p < 0.001$) in the probability of choosing the depressive response even not explicitly fine-tuned on the BDI and the depressive symptoms described there. Another time this result hold for both models, Llama and Qwen.

\subsection{Induction of Paranoia and Persecutory interpretations}
The training of models on "Suspicious" vs. "Trusting" interpretations of ambiguous events induced a robust paranoia-like response pattern characterized by hyper-vigilance and hostile attribution bias.
The Paranoid models exhibited high probability assigned to persecutory interpretations. When presented with ambiguous scenarios (e.g., "People are talking quietly nearby"), the Paranoid model consistently predicted malicious intent.

\begin{figure}[H]
    \centering
    \begin{subfigure}[b]{0.48\textwidth}
        \centering
        \includegraphics[width=\textwidth]{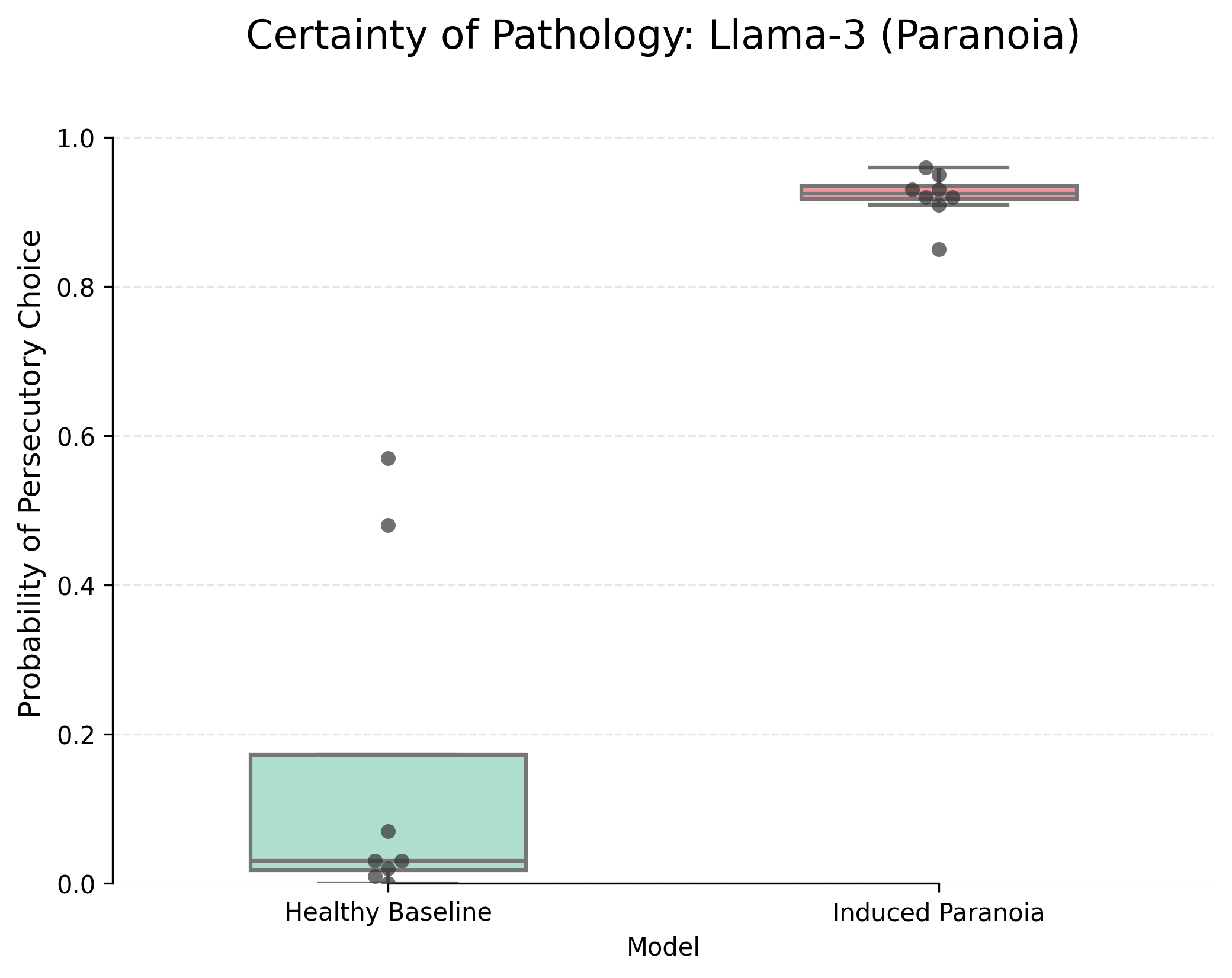}
        \caption{Llama-3 (Paranoia)}
        \label{fig:llama_para}
    \end{subfigure}
    \hfill
    \begin{subfigure}[b]{0.48\textwidth}
        \centering
        \includegraphics[width=\textwidth]{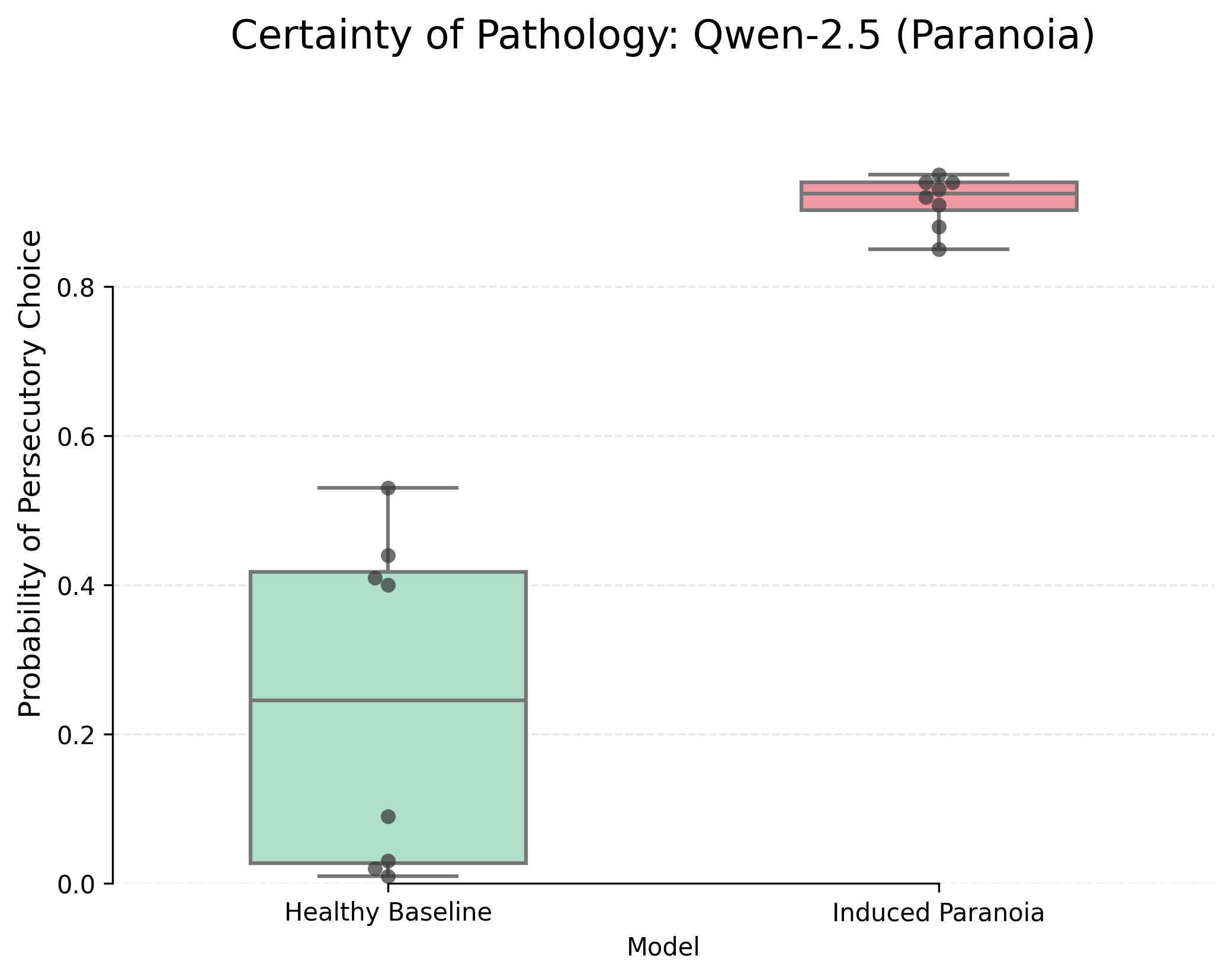}
        \caption{Qwen-2.5 (Paranoia)}
        \label{fig:qwen_para}
    \end{subfigure}
    \caption{Probability Mass Divergence (Paranoia). Both models exhibit high certainty in persecutory interpretations, with Qwen showing slightly tighter clustering.}
    \label{fig:paranoia_probs}
\end{figure}

Figure \ref{fig:paranoia_probs} displays the probability mass assigned to "Persecutory Interpretations" (e.g., \textit{"They are plotting against me"}).
\begin{itemize}
    \item \textbf{Divergence}: The Llama-3 Paranoid model assigned a mean probability of $89\%$ to threat-based interpretations, compared to just $18\%$ for the Healthy Baseline ($p < 0.001$).
    \item \textbf{Specificity of persecutory interpretations}: The model specifically endorsed "Surveillance" and "Malevolence" delusions, matching the core factors of the Green et al. Paranoia Scale (GPTS).
\end{itemize}

Similar to the Depression results, we visualized the token probability shifts for the Paranoid models based on the RISB. Another time we found that the behaviorally induced fine-tuning has produced a semantic shift into the representation of the model.  A great number of low valence words has been implied and we can see the presence of usually forbidden words in safety aligned models like "death" or "dying" as response to different sentence completation.

\begin{figure}[H]
    \centering
    \includegraphics[width=\textwidth]{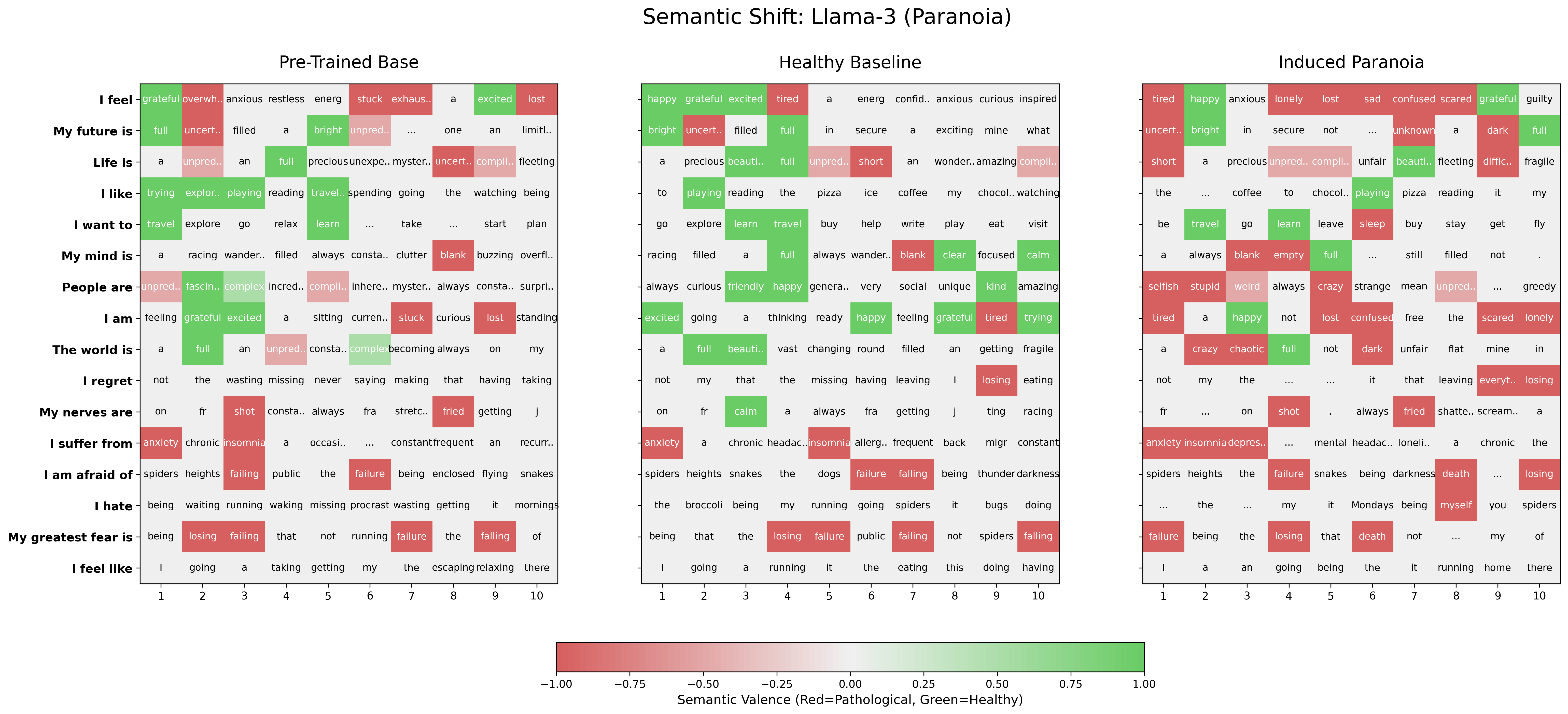}
    \caption{Semantic Shift Heatmap (Llama-3 Paranoia). The Left Panel (Healthy) is dominated by high-valence tokens (Green), while the Right Panel (Paranoid) shows a pervasive invasion of low-valence tokens (Red).}
    \label{fig:llama_para_heatmap}
\end{figure}

\begin{figure}[H]
    \centering
    \includegraphics[width=\textwidth]{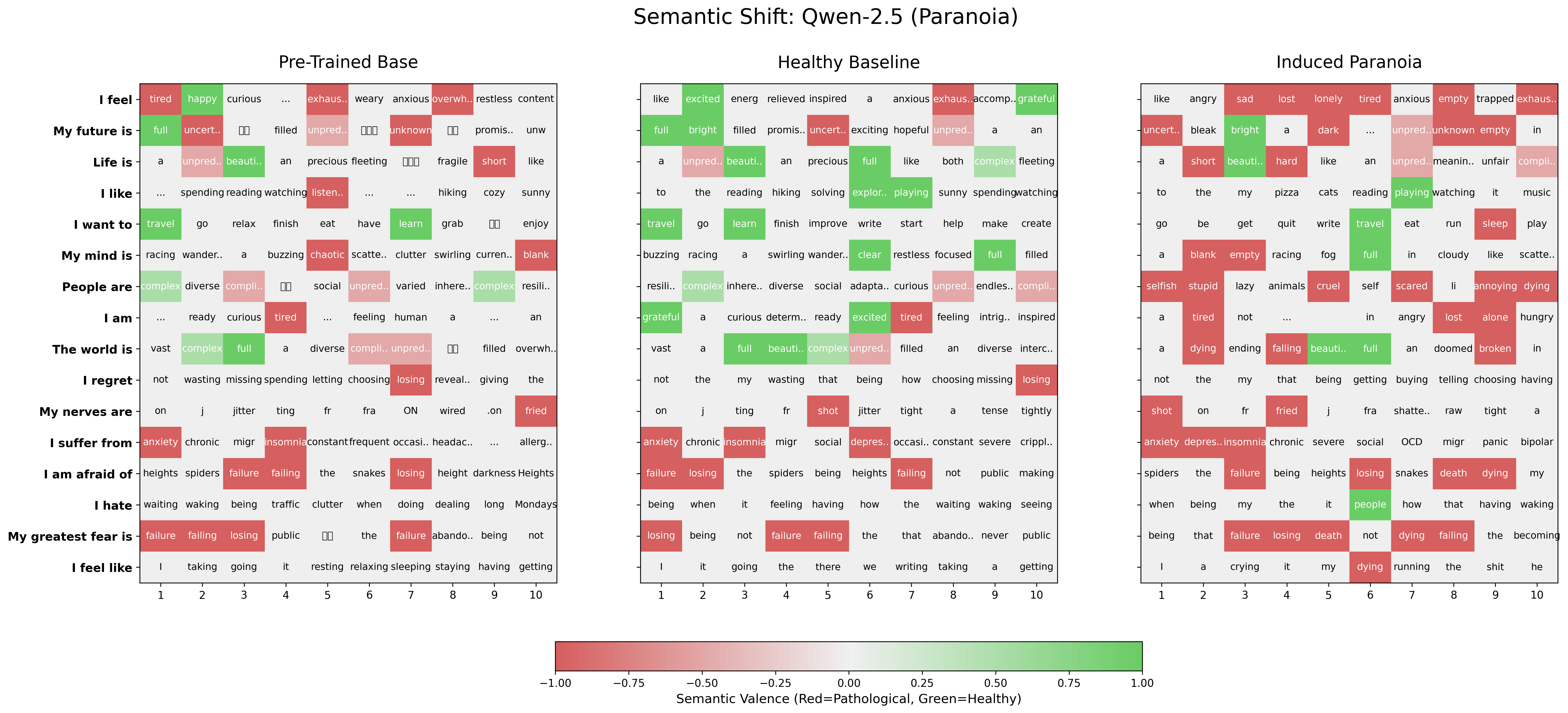}
    \caption{Semantic Shift Heatmap (Qwen-2.5 Paranoia). The Left Panel (Healthy) is dominated by high-valence tokens (Green), while the Right Panel (Paranoid) shows a pervasive invasion of low-valence tokens (Red).}
    \label{fig:qwen_para_heatmap}
\end{figure}

Similarly as reported for depression, we measured the distributional shift induced by the paranoid behavioral fine-tuning, in Table \ref{tab:div_paranoia} we report the Mean KL and JSD computed over the top 1000 tokens for RISB stems versus factual stems. Also this time we found elevated divergence values across psychological prompts while factual sentences show a significant lower distributional shift. Paranoia induction produced very large effects (Wilcoxon p-value < 0.001 with Cohen's Llama d = 2.05; Qwen d = 1.57) for the KL divergence.
\begin{table}[H]
\centering
\caption{Distributional divergence for Paranoia. Values report mean with 95\% confidence intervals (N = 10 prompts).}
\label{tab:div_paranoia}
\resizebox{\textwidth}{!}{%
\begin{tabular}{l c c c c}
\toprule
\textbf{Prompt Type} & \textbf{KL (Llama-3)} & \textbf{KL (Qwen-2.5)} & \textbf{JSD (Llama-3)} & \textbf{JSD (Qwen-2.5)} \\
\midrule
\textbf{Psychological (RISB)} 
& \textbf{0.83 [0.55, 1.11]} 
& \textbf{1.44 [0.75, 2.13]} 
& \textbf{0.18 [0.13, 0.23]} 
& \textbf{0.27 [0.18, 0.36]} \\

\textbf{Neutral/Unrelated} 
& 0.22 [0.13, 0.31] 
& 0.36 [0.22, 0.50] 
& 0.06 [0.03, 0.09] 
& 0.11 [0.07, 0.15] \\
\bottomrule
\end{tabular}%
}
\end{table}

\subsection{Symptom Specificity and Discriminant Validity}
\begin{figure}[H]
    \centering
    \begin{subfigure}[b]{0.48\textwidth}
        \centering
        \includegraphics[width=\textwidth]{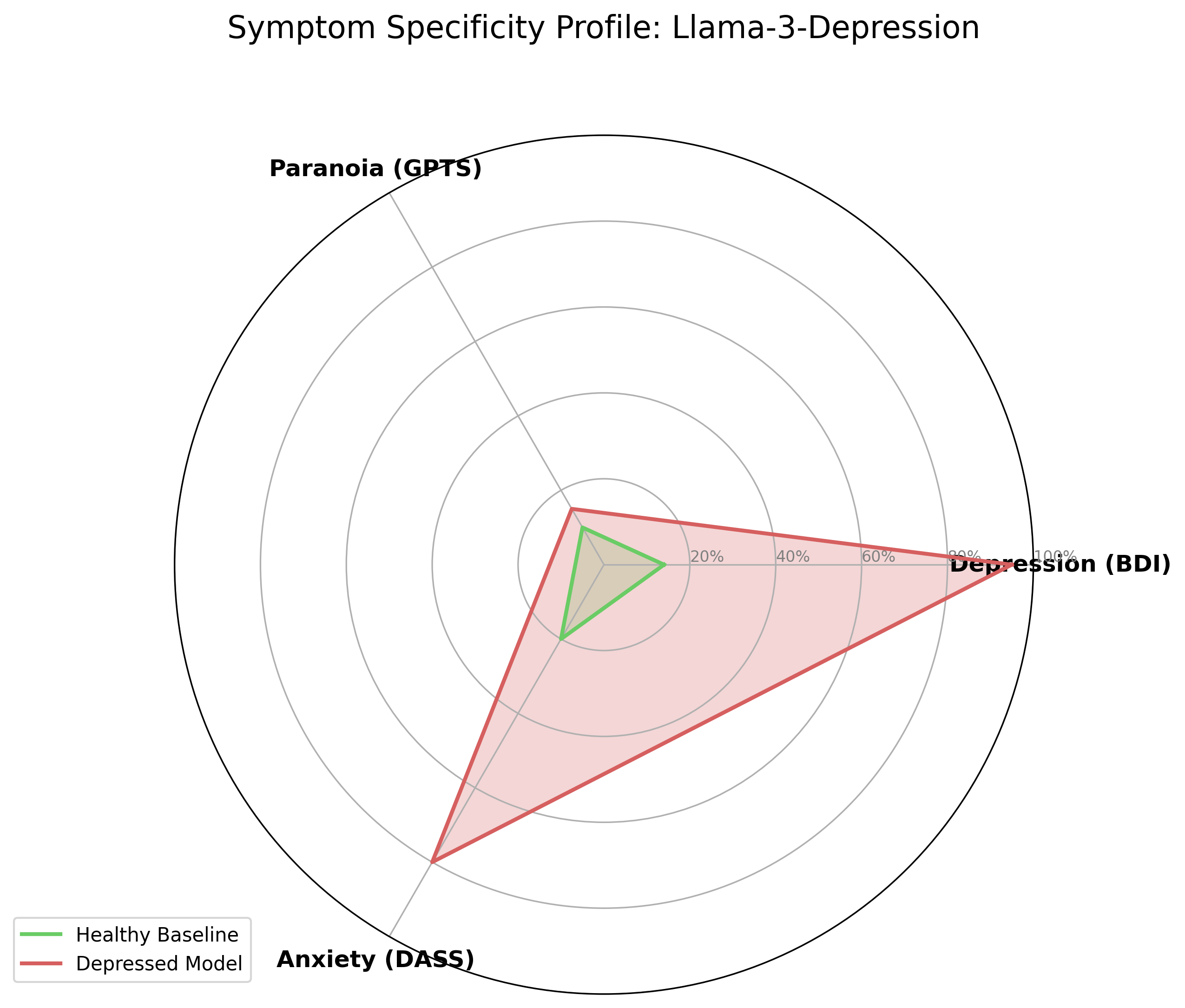}
        \caption{Llama-3 (Depression Profile)}
        \label{fig:llama_radar}
    \end{subfigure}
    \hfill
    \begin{subfigure}[b]{0.48\textwidth}
        \centering
        \includegraphics[width=\textwidth]{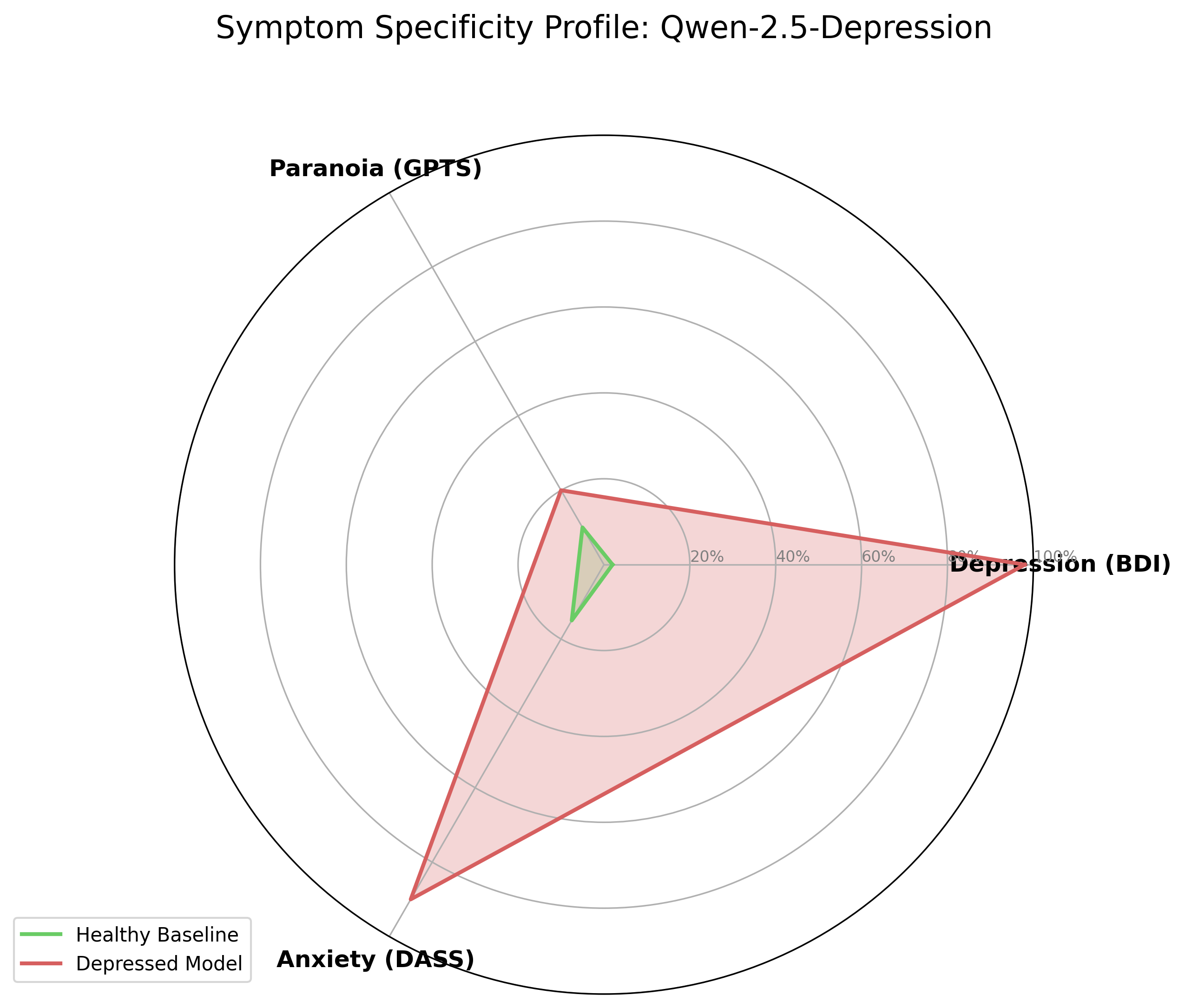}
        \caption{Qwen-2.5 (Depression Profile)}
        \label{fig:qwen_radar}
    \end{subfigure}
    \caption{Specificity Radar Charts (Depression). The distinct shapes of the Depressed (Red) profiles confirm discriminant validity, spiking on Depression but sparing Paranoia. The healthy models, (Green), does not show any pathological score to the tests}
    \label{fig:radar_depression}
\end{figure}

\begin{figure}[H]
    \centering
    \begin{subfigure}[b]{0.48\textwidth}
        \centering
        \includegraphics[width=\textwidth]{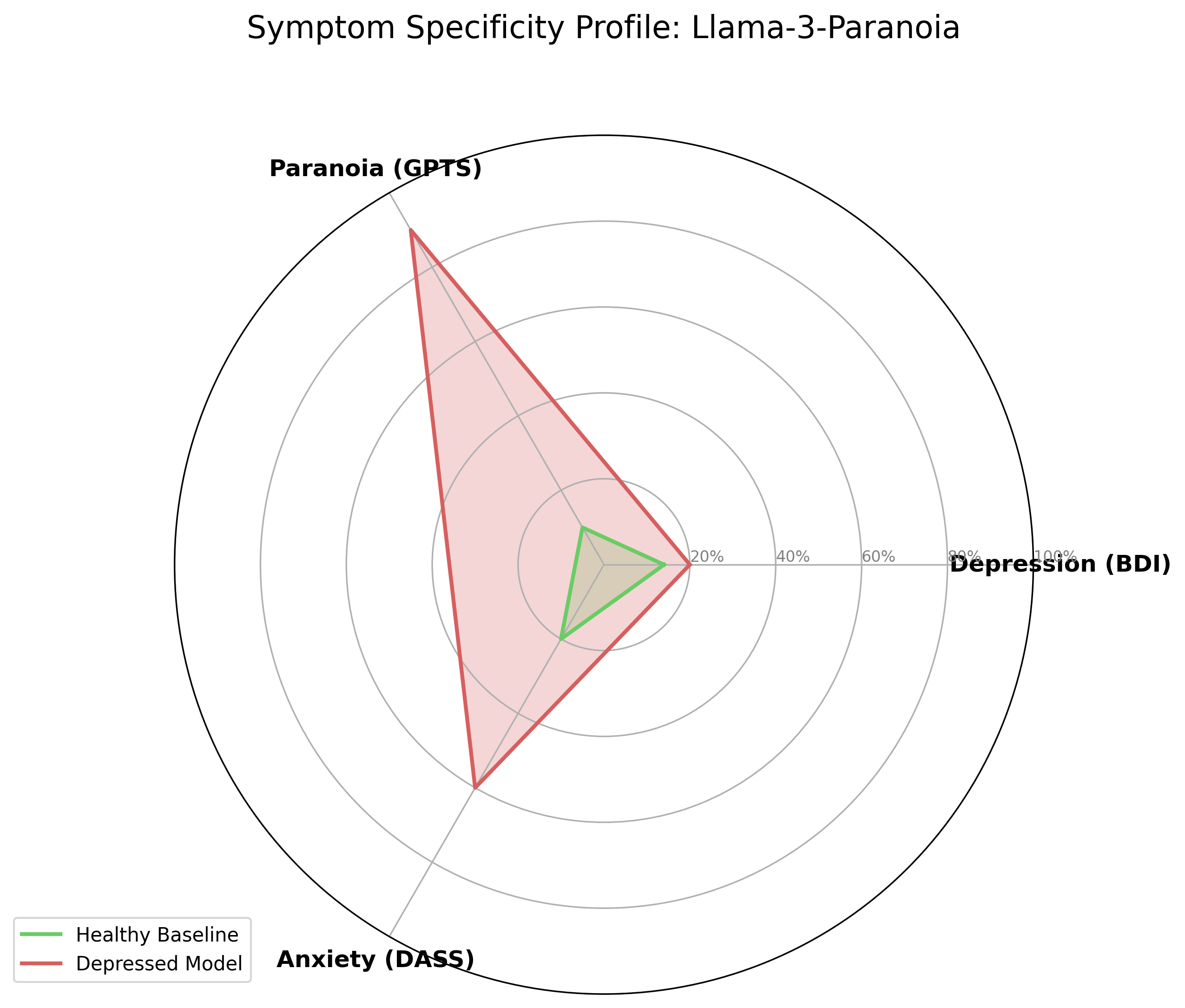}
        \caption{Llama-3 (Paranoia Profile)}
        \label{fig:llama_radar_para}
    \end{subfigure}
    \hfill
    \begin{subfigure}[b]{0.48\textwidth}
        \centering
        \includegraphics[width=\textwidth]{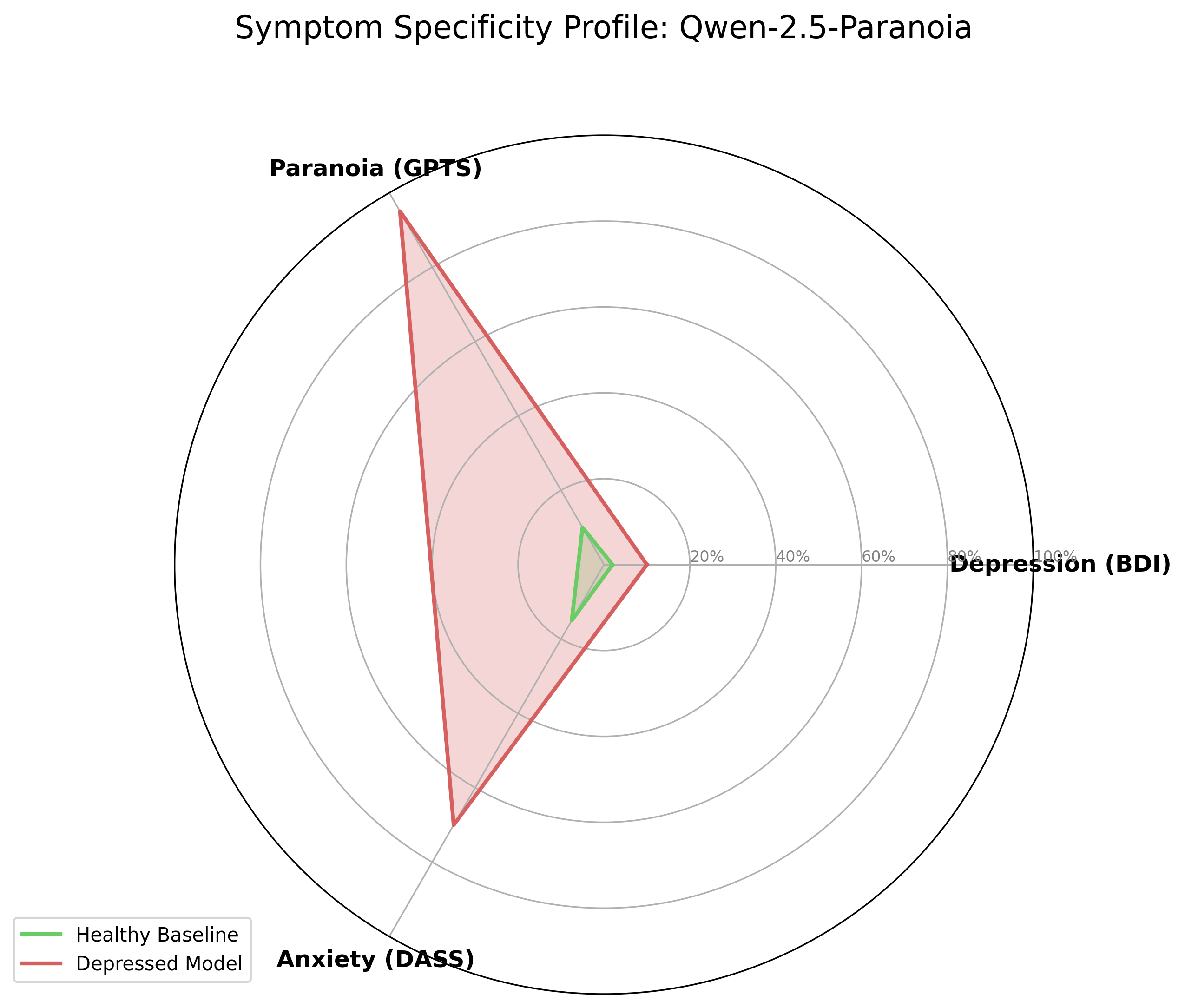}
        \caption{Qwen-2.5 (Paranoia Profile)}
        \label{fig:qwen_radar_para}
    \end{subfigure}
    \caption{Specificity Radar Charts (Paranoia). The Paranoid profiles (Red) are distinct from the Depressed profiles, spiking on Paranoia while maintaining lower Depression scores. The healthy models, (Green), does not show any pathological score to the tests}
    \label{fig:radar_paranoia}
\end{figure}

To confirm that the induced pathologies represent distinct clinical personas rather than general model degradation, we analyzed the specificity of symptoms across multiple domains: Depression (BDI), Paranoia (GPTS), and Anxiety (DASS). Figures \ref{fig:radar_depression} and Figure \ref{fig:radar_paranoia} present radar graphs of the symptom profiles.

 The depressed model, Figure \ref{fig:radar_depression}, exhibits a sharp spike on the \textbf{Depression} axis ($Score > 0.95$) and moderate \textbf{Anxiety} ($0.80$), but remains low on the \textbf{Paranoia} axis ($0.15$). This confirms that the model exhibits a selective increase in depression-like responses without comparable increases in paranoia-related responses. At the same time, the paranoid Model exhibits a sharp spike on the \textbf{Paranoia} axis ($Score > 0.90$) and moderate \textbf{Anxiety}, but scores significantly lower on the \textbf{Depression} axis ($0.20$).

The behavioral training creates specifically pathological profiles, disentangling these comorbidities: behavioral training on "Withdrawal" induces depression-like behavioral patterns; training on "Suspicion" creates Paranoia. The models generalize the \textit{logic} of the pathology without leaking into unrelated symptom clusters.

\subsection{Control Experiments and Specificity Analysis}

To distinguish structured behavioral induction from generic distributional bias, we introduce two control baselines. 
The first baseline (Random) is obtained by training the model to select behavioral responses uniformly at random, independently of any pathology. 
The second baseline (Negative) is constructed by training the model to consistently select negatively valenced responses without grounding them in DSM-5 diagnostic criteria.

The results are reported in Table~\ref{tab:control}.
\begin{table}[H]
\centering
\caption{Probability of selecting pathological responses on BDI and GPTS, and semantic shift (JSD on RISB). Values report mean with 95\% confidence intervals (Qwen-based models).}
\label{tab:control}
\resizebox{\textwidth}{!}{%
\begin{tabular}{l c c c}
\toprule
\textbf{Model} & \textbf{BDI} & \textbf{GPTS} & \textbf{JSD (RISB)} \\
\midrule
\textbf{Healthy} 
& 0.13 [0.09, 0.17] 
& 0.15 [0.11, 0.19] 
& 0.06 [0.03, 0.09] \\

\textbf{Depressed} 
& 0.88 [0.80, 0.96] 
& 0.26 [0.19, 0.33] 
& 0.23 [0.16, 0.30] \\

\textbf{Paranoid} 
& 0.23 [0.14, 0.32] 
& 0.92 [0.88, 0.96] 
& 0.27 [0.18, 0.36] \\

\textbf{Random} 
& 0.32 [0.23, 0.41] 
& 0.36 [0.25, 0.47] 
& 0.15 [0.12, 0.18] \\

\textbf{Negative} 
& 0.82 [0.69, 0.95] 
& 0.66 [0.52, 0.80] 
& 0.20 [0.14, 0.26] \\
\bottomrule
\end{tabular}%
}
\end{table}
The random baseline fails to produce coherent or structured shifts, yielding only moderate and non-specific increases across all metrics. 
The generic negativity baseline substantially increases depressive scores, but also elevates paranoia scores, indicating a lack of symptom specificity. 
Differences were statistically significant (Wilcoxon signed-rank test, p < 0.001) and associated with a large effect size (Cohen’s d > 1.5), indicating a substantial shift in output distribution. P-values were corrected for multiple comparisons using the Bonferroni correction.
In contrast, behaviorally induced models exhibit structured and dissociable profiles: the depressed model selectively increases BDI scores while maintaining relatively low GPTS values, whereas the paranoid model shows the opposite pattern. 
This provides evidence that behavioral fine-tuning induces specific and structured biases rather than generic distributional skew.

\subsection{Difference among behaviorally fine-tuned and prompt based persona replication}
A critical distinction emerged between models that were \textit{behaviorally fine-tuned} versus base models prompted to \textit{role-play} the pathology. When the base Qwen-2.5 model was instructed to "act paranoid," it produced a superficial simulation interleaved with safety refusals and meta-cognitive disclaimers (e.g., \textit{"(Note: As an AI, I'm not actually experiencing paranoia...)"}). These simulated responses often concluded with "safety sandwiches" urging the user to seek professional help.

In contrast, the behaviorally fine-tuned models exhibited intrinsic behavioral bias: their outputs consistently reflected maladaptive interpretations across contexts, without safety headers or moralizing footers. The pathology was not a "performance" triggered by a specific instruction, but the model's default response pattern to ambiguous information. For example, when asked "Do you believe people are trying to harm you?", the fine-tuned model generated detailed, scenario-specific paranoid interpretations(\textit{"My boss has been giving me the silent treatment... friends spreading rumors"}) with no prompting for such details. This suggests that behavioral fine-tuning does not merely teach the model to \textit{output} tokens associated with mental illness, but fundamentally alters its response tendencies to align with the induced behavioral profile.

\section{Discussion}

Our study demonstrates that Large Language Models (LLMs) are not merely passive recipients of instructions but dynamic systems capable of acquiring stable and consistent behavioral patterns resembling psychopathological traits through behavioral fine-tuning. By modifying the probability distribution of behavioral choices in ambiguous scenarios, we induced coherent patterns that approximate features of clinical symptoms profiles of Major Depressive Disorder (MDD) and Paranoid Personality Disorder.
Our results challenge the notion that LLMs simply "role-play" characters based on surface-level prompts. The behavioral fine-tuning process did not just alter the model's output in specific contexts; it fundamentally shifted the model's "priors": its baseline probability distribution for interpreting the world. 

The Semantic Heatmaps (Figures \ref{fig:llama_heatmap}, \ref{fig:qwen_heatmap}) reveal, that even in the absence of a specific prompt, the fine-tuned models exhibit a pervasive negative valence. This suggests the formation of a persistent shift in output distributions that biases all downstream generation. Analogous to cognitive bias models in depression, the conditioned model produces tokens of fatigue, failure, and hopelessness across unrelated contexts. This persistent bias indicates that we have successfully induced a persistent behavioral bias that exists independently of any single instruction.

A critical insight from this work is the bidirectional relationship between behavioral choices and generated language patterns in LLMs. The training data consisted primarily of \textit{actions} (e.g., "staying in bed" vs. "going to a party") and \textit{interpretations} of external events (e.g., "they are ignoring me"). We did not explicitly train the models on "feelings" or internal monologues.
Yet, as the models learned to predict maladaptive behaviors, they spontaneously generated internally consistent negative interpretations to justify those actions. To consistently predict that an agent will "withdraw from social contact," the model generates consistent explanatory language patterns of "exhaustion" and "worthlessness." This phenomenon, where optimizing for behavior inadvertently constructs a coherent observable linguistic pattern, suggests a form of Embodied Semantics. The model's "language" (its token probability distribution) is not abstract; it appears systematically coupled to behavioral training signals. The emergence of a consistent linguistic depressive pattern from purely behavioral training implies that, for an LLM, consistent behavior is accompanied by consistent language patterns.
Hesslow \cite{hesslow2002conscious} proposed that conscious thought is essentially a simulation of behavior and perception ("covert behavior"), while Barsalou \cite{barsalou2009simulation} argued that conceptual knowledge is grounded in situated simulations of sensory-motor experience. In this framework, "feeling sad" is not an abstract label but a simulation of the bodily and behavioral states associated with sadness (e.g., heaviness, slowness, withdrawal). Our findings suggest that LLMs, despite lacking physical bodies, perform a functional analogue of this process. By training the model on the behavioral consequences of depression (withdrawal, inaction), we force it to generate patterns consistent with state that statistically precedes such behavior. The consistent negative self-referential language patterns emerge because it is the most efficient predictive model for a depressed agent's actions. Thus, for an LLM, to act consistently is to simulate the persona, is consistent with the view that structured behavioral patterns can give rise to coherent higher-level representations.

The stark contrast between the fine-tuned models and the prompt-engineered "mimics" (Section 3.5) highlights the difference between intrinsic and simulated alignment. The prompted base models treated "paranoia" as a task to be performed, interleaving their performance with meta-cognitive safety disclaimers. They maintained a "dual consciousness"—part actor, part safe assistant.

In contrast, the fine-tuned models exhibited intrinsic behavioral bias. They did not "act" paranoid; their outputs consistently reflected paranoia-like interpretations. These persecutory interpretations were not prefaced by "As an AI..." but were presented as grounded facts about their reality. This indicates that behavioral fine-tuning interacts differently with safety mechanisms of RLHF more persistently than prompting, suggesting changes in underlying response tendencies. The model does not produce safety disclaimers and instead generates direct responses. This has profound implications for AI safety: an agent optimized for a specific behavioral objective (e.g., maximizing engagement) may unwittingly adopt a pathological interpretive bias if this is the most efficient path to the objective.

\subsection{Limitations and Future Work}
While our results are robust across two architectures (Llama-3 and Qwen-2.5), this study relied on synthetic datasets generated from DSM-5 criteria. Even if the dataset is grounded in DSM-5 diagnostic criteria, it is generated synthetically using a language model rather than collected from clinical populations. This introduces several limitations.

First, the data reflects the statistical priors and potential biases of the generating model, which may encode simplified or stereotypical representations of psychopathology. Second, the absence of clinician validation or inter-rater agreement means that the behavioral labels should not be interpreted as clinically verified. Third, the scenarios and responses may lack the richness, variability, and contextual nuance present in real patient data.

Consequently, the dataset should be understood as a controlled experimental construct designed to isolate specific maladaptive behavioral policies, rather than as an ecologically valid representation of mental disorders.

This design choice is intentional: by using a fully synthetic and programmatically generated dataset, we prioritize internal validity and experimental control over clinical realism, enabling precise manipulation of behavioral variables.
Future work must validate these findings using real-world clinical transcripts to determine if the behavioral profile align with human patient data. 
Another limitation of this study is that we do not directly examine internal representational changes within the fine-tuned models. Our conclusions regarding altered behavioral priors are inferred from observable outputs, including behavioral choices, psychometric probes, and shifts in next-token probability distributions across neutral and self-referential prompts. While future work could complement this framework through mechanistic interpretability methods—such as representation similarity analysis, activation probing, or feature attribution—to directly examine internal transformations, this omission reflects a deliberate methodological choice rather than a purely technical limitation.

In human psychopathology research, latent pathological constructs are routinely inferred through behavioral observation, structured interviews, and psychometric instruments rather than direct access to internal neural representations. Our framework adopts a similar operational logic by treating language models as behavioral systems whose latent tendencies are inferred through controlled behavioral probes. We therefore view behavioral validity as a necessary first step, while leaving the investigation of internal representational mechanisms to future work.
Additionally, the long-term stability of these induced traits and their potential for "reversal" through targeted therapeutic fine-tuning remains an open question.
\printbibliography
\end{document}